\documentclass[journal]{IEEEtran}
\usepackage{cite}
\usepackage{graphicx}
\usepackage{amsmath}
\usepackage{amsfonts}
\usepackage{amssymb}
\usepackage{algorithmic}
\usepackage{array}
\usepackage{url}
\usepackage{multirow}
\ifCLASSOPTIONcompsoc
    \usepackage[caption=false, font=normalsize, labelfont=sf, textfont=sf]{subfig}
\else
\usepackage[caption=false, font=footnotesize]{subfig}
\fi

\begin{document}
%
\title{GAN-based Synthetic Medical Image Augmentation  \\ for increased CNN Performance \\ in Liver Lesion Classification}
%
%
\author{Maayan~Frid-Adar, Idit~Diamant, Eyal~Klang, Michal~Amitai,
        Jacob~Goldberger,     		   and~Hayit~Greenspan,~\IEEEmembership{Member,~IEEE}
        
\thanks{M. Frid-Adar, I. Diamant and H. Greenspan are with the Department
of Biomedical Engineering, Tel Aviv University, Tel Aviv, Israel
(e-mail: maayan.frid@gmail.com; iditdiamant@gmail.com; hayit@eng.tau.ac.il).}
\thanks{E. Klang and M. Amitai are with the Department of Diagnostic Imaging,
The Chaim Sheba Medical Center, Tel-Hashomer, Israel (e-mail:
eyalkla@hotmail.com; michal.amitai@sheba.health.gov.il).}
\thanks{J. Goldberger is with the Faculty of Engineering, Bar-Ilan University, Ramat-Gan, Israel (e-mail: jacob.goldberger@biu.ac.il).}}



\maketitle

\begin{abstract}
Deep learning methods, and in particular convolutional neural networks (CNNs), have led to an enormous breakthrough in a wide range of computer vision tasks, primarily by using large-scale annotated datasets.
However, obtaining such datasets in the medical domain remains a challenge.
In this paper, we present methods for generating synthetic medical images using recently presented deep learning Generative Adversarial Networks (GANs).
Furthermore, we show that generated medical images can be used for synthetic data augmentation, and improve the performance of CNN for medical image classification.
Our novel method is demonstrated on a limited dataset of computed tomography (CT) images of 182 liver lesions (53 cysts, 64 metastases and 65 hemangiomas). We first exploit GAN architectures for synthesizing high quality liver lesion ROIs. Then we present a novel scheme for liver lesion classification using CNN. Finally, we train the CNN using classic data augmentation and our synthetic data augmentation and compare performance. In addition, we explore the quality of our synthesized examples using visualization and expert assessment.
The classification performance using only classic data augmentation yielded 78.6\% sensitivity and 88.4\% specificity. By adding the synthetic data augmentation the results increased to 85.7\% sensitivity and 92.4\% specificity.
We believe that this approach to synthetic data augmentation can generalize to other medical classification applications and thus support radiologists' efforts to improve diagnosis.

\end{abstract}



\section{Introduction} \label{sec:introduction}

\IEEEPARstart{T}{he} greatest challenge in the medical imaging domain is how to cope with the small datasets and limited amount of annotated samples
\cite{Roth2016,Litjens2017deepMedicalSurvey,Greenspan2016overviewDeepMedical,Tajbakhsh2016CNNmedical,Shi2016}, especially when employing supervised machine learning algorithms that require labeled data and larger training examples.
In medical imaging tasks, annotations are made by radiologists with expert knowledge on the data and task. Most annotations of medical images are time consuming. This is especially true for  precise annotations, such as the segmentations of organs or lesions into multiple 2-D slices and 3-D volumes. 
Although public medical datasets are available online, and grand challenges have been publicized, 
most datasets are still limited in size and only applicable to specific medical problems.
Collecting medical data is a complex and expensive procedure that requires the collaboration of researchers and radiologists \cite{Greenspan2016overviewDeepMedical}.

Researchers attempt to overcome this challenge by using data augmentation. The most common data augmentation methods include simple modifications of dataset images such as translation, rotation, flip and scale. Using classic data augmentation to improve the training process of networks is a standard procedure in computer vision tasks \cite{Alexnet}. However, little additional information can be gained from small modifications to the images (e.g. the translation of the image a few pixels to the right).
Synthetic data augmentation of high quality examples is new, sophisticated type of data augmentation. Synthetic data examples learned using a generative model enable more variability and enrich the dataset to further improve the system training process.

One such promising approach inspired by game theory for training a model that syntheses images is known as Generative Adversarial Networks (GANs) \cite{Goodfellow2014GAN}. 
The model consists of two networks that are trained in an adversarial process where one network generates fake images and the other network discriminates between real and fake images repeatedly.
GANs have gained great popularity in the computer vision community and different variations of GANs were recently proposed for generating high quality realistic natural images \cite{Radford2015DCGAN,Denton2015lapgan,Mirza2014conditionalGAN,Odena2016ACGAN}. Interesting applications of GAN include generating images of one style from another style (image-to-image translation) \cite{Isola2016imageToimage} and image inpainting using GAN \cite{Yeh2016GANinpainting}.

Recently, several medical imaging applications have applied the GAN framework \cite{costa2017towards,dai2017scan,xue2017segan,nie2016medical,Schlegl2017ganAnomaly,alex2017generative,Ben_Cohen2017PET}. Most studies have employed the image-to-image GAN technique to create label-to-segmentation translation, segmentation-to-image translation or medical cross modality translations.
Costa et al. \cite{costa2017towards} trained a fully-convolutional network to learn retinal vessel segmentation images. Then they learned the translation from the binary vessel tree to a new retinal image. Dai et al. \cite{dai2017scan} trained GAN to create segmentation images of the lung fields and the heart from chest X-ray images.
Xue et al. \cite{xue2017segan} referred to the two GAN networks as a Segmentor and Critic, and learned the translation between brain MRI images and a brain tumor binary segmentation map. 
In Nie et al. \cite{nie2016medical}, A patch-based GAN was trained for translation between brain CT images and the corresponding MRI images. They further suggested an auto-context model for image refinement. 
Ben-Cohen et al. \cite{Ben_Cohen2017PET} also introduced a cross modality image generation using GAN, from abdominal CT image to a PET scan image that highlights liver lesions.
Some studies have been inspired by the GAN method for image inpainting. 
Schlegl et al. \cite{Schlegl2017ganAnomaly} trained GAN with healthy patches of the retinal area to learn the data distribution of healthy tissue. Then they tested the GAN on patches of both unseen healthy and anomalous data for anomaly detection in retinal images. 

The problem of limited data in the medical imaging field prompted us to explore methods for synthetic data augmentation to enlarge medical datasets. In the current study, we focus on improving results in the specific task of liver lesion classification. We applied the GAN framework to synthesize high quality liver lesion images (hereon we use interchangeably the terms lesion images and lesion ROIs). 

The liver is one of three most common sites for metastatic cancer along with the bone and lungs \cite{metastases_info}. According to the World Health Organization, in 2012 alone, cancer accounted for 8.2 million deaths worldwide of which 745,000 were caused by liver cancer \cite{Ferlay2015cancer}. Focal liver lesions can be malignant and manifest metastases, or be benign (e.g. hemangioma or hepatic cysts). Computed tomography (CT) is one of the most common and robust imaging techniques for the detection, diagnosis and follow up of liver lesions \cite{Murakami2011CT}.
Thus, there is a great need and interest in developing automated diagnostic tools based on CT images to assists radiologists in the diagnosis of liver lesions.

Previous studies have presented methods for automatic classification of focal liver lesions in CT images \cite{Gletsos2003,Adcock2014,Chang2017,Bilello2004,Mougiakakou2007,Diamant2016improvedClassify,Diamant2017BOVWMI}.
Gletsos et al. \cite{Gletsos2003} used texture features for liver lesion classification into four categories including the normal liver parenchyma class. They applied a hierarchical classifier of neural networks at each level. 
Chang et al. \cite{Chang2017} obtained three kind
of features for each tumor, including texture, shape, and kinetic curve on segmented tumors. Backward elimination was used to select the best combination of features through binary logistic regression analysis to classify the tumors.
Diamant et al. \cite{Diamant2016improvedClassify} applied the bag-of-visual-words (BoVW) method learned from image patches. They used two dictionaries for lesion interior and boundary regions. Based on the two dictionaries they generated histograms for each lesion ROI. The final classification was made using SVM.

In the current work we used deep learning methodology for the task of liver lesion classification. 
Deep learning convolutional neural networks (CNNs) has emerged as a powerful tool in computer vision. In recent years many medical imaging studies have applied CNNs and reported improved performance for a broad range of medical tasks \cite{Greenspan2016overviewDeepMedical}.
We combine synthetic liver lesion generation using GAN with our proposed CNN for liver lesion classification.

The contributions of this work are the following:
\begin{enumerate}
\item Synthesis of high quality focal liver lesions from CT images using generative adversarial networks (GANs).
\item Design of a CNN-based solution for the liver lesion classification task, with comparable results to state-of-the-art methods.
\item Augmentation of the CNN training set, using the generated synthetic data - for improved classification results. 
\end{enumerate}

\section{Liver Lesion Classification} \label{sec:data_classification}

In this section we first describe the data and their characteristics. Then we elaborate on the CNN architecture for the liver lesion classification task. 
The main challenge is the small amount of data available for training the CNN. In the next section we describe methods to artificially enlarge the data.

\subsection{Data} \label{sec:data}
The dataset used in this work contains cases of liver lesions collected from Sheba Medical Center by searching medical records for cases of cysts, metastases and hemangiomas. Cases were acquired from 2009 to 2014 using two CT scanners: a General Electric (GE) Healthcare scanner and a Siemens Medical System scanner, with the following parameters: 120kVp, 140-400mAs and 1.25-5.0 mm slice thickness.
Cases were collected with the approval of the institution's Institutional Review Board.

\begin{figure*}[!t]
\centering
\subfloat[]{\includegraphics[width=4.0in]{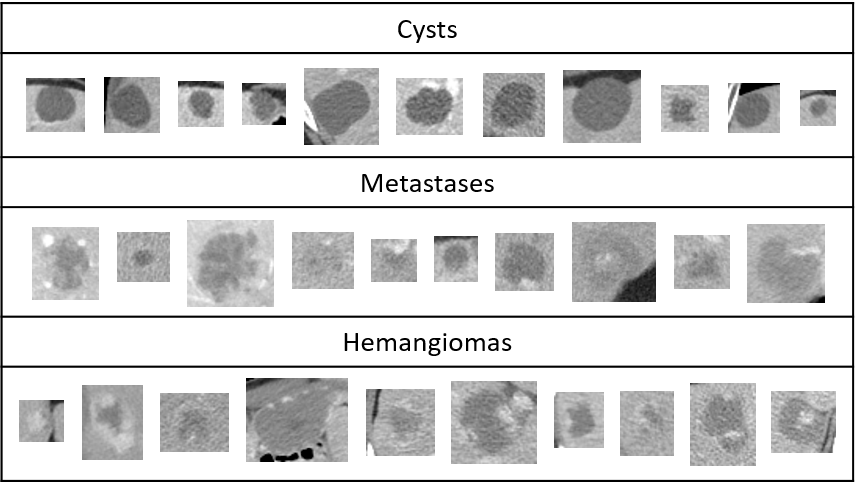}%
\label{fig:lesions_example}}
\hfil
\subfloat[]{\includegraphics[width=1.8in]{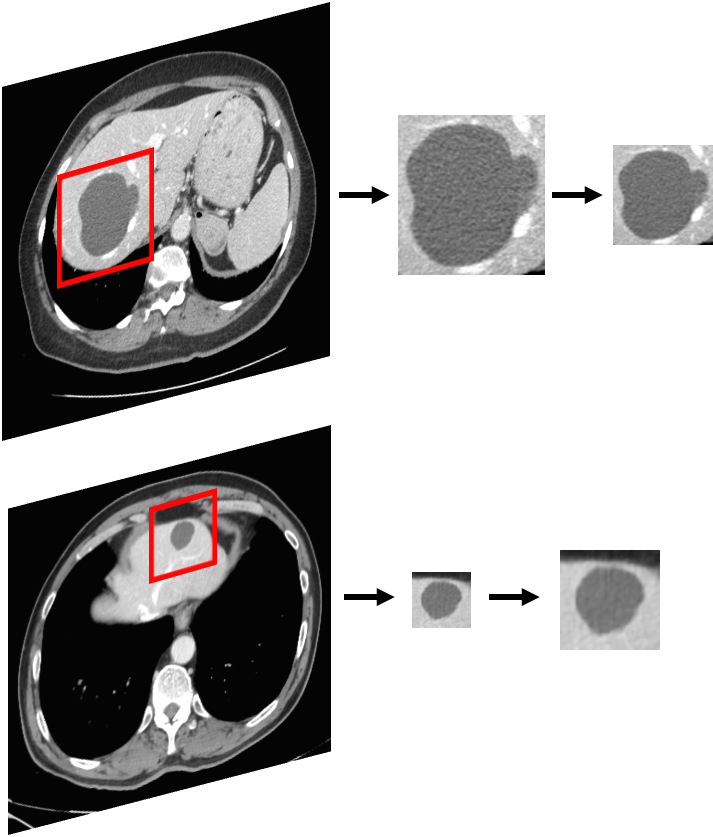}%
\label{fig:roi_extraction_two}}
\caption{(a) Dataset examples of cyst, metastasis and hemangioma liver lesions. (b) ROI extraction process from a 2-D CT slice of the liver. All ROIs are resized to a uniform size.}
\label{fig:two_figure_lesion_examples}
\end{figure*}

Figure \ref{fig:two_figure_lesion_examples} shows examples of the input data and the ROI extraction process. The dataset  was made up of 182 portal-phase 2-D CT scans (Figure \ref{fig:lesions_example}): 53 cysts, 64 metastases, 65 hemangiomas. An expert radiologist marked the margin of each lesion and determined its corresponding diagnosis which was established by biopsy or a clinical follow-up. This serves as our ground truth.

Liver lesions vary considerably in shape, contrast and size (10 - 102mm). They also vary within categories. In addition, they are located in interior sections of the liver or near its boundary where the surrounding parenchyma tissue of the lesions changes.
Each type of lesion has its own characteristics: Cysts are non-enhancing water-attenuation circumscribed lesions. Metastases are hypoattenuating, have soft-tissue attenuation and less well-defined margins than cysts, and hemangiomas show typical features of discontinuous nodular peripheral enhancement, with fill-in on delayed images \cite{Napel2010}.
Despite this detailed description, some characteristics may be confusing, in particular for metastasis and hemangioma lesions (see Figure \ref{fig:lesions_example}). Metastases can contain areas of higher density, probably prominent blood vessels or calcifications that can be mistaken for hemangiomas attributes.
Hemangiomas are benign tumors and metastases are malignant lesions derived from different primary cancers. Thus, the correct identification of a lesion as metastasis or hemangioma is especially important.

The input to our classification system are ROIs of lesions cropped from CT scans using the radiologist's annotations.  The ROIs are extracted to capture the lesion and its surrounding tissue relative to its size. Due to the large variability in lesion sizes, this results in varying size ROIs (Figure \ref{fig:roi_extraction_two}).

\subsection{CNN Architecture} \label{sec:classification_cnn}
\begin{figure}[!t]
\centering
\includegraphics[width=1.7in]{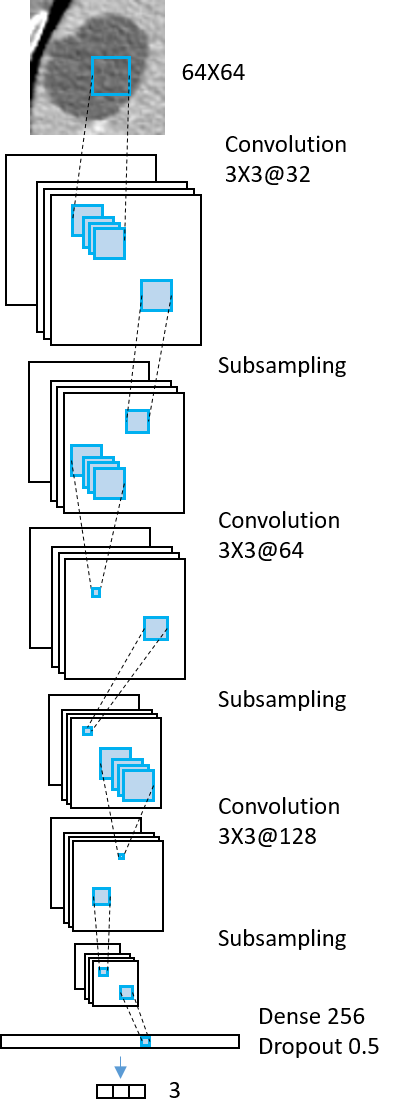}
\caption{The architecture of the liver lesion classification CNN.}
\label{fig:cnn_arch}
\end{figure}

The architecture of the liver lesion classification system we propose 
is shown in Figure \ref{fig:cnn_arch}. CNNs are widely used for solving image classification tasks in computer vision \cite{Alexnet}. CNN architectures for medical imaging have also been introduced \cite{Shin2016,Roth2016,Setio2016}, usually containing fewer convolutional layers because of the small datasets and smaller input size. Our classification CNN gets fixed size input ROIs of $64\times{64}$,  with an intensity range rescaled to $(0,1)$. The architecture consists of three pairs of convolutional layers where each convolutional layer is followed by a max-pooling layer, and two dense fully-connected layers ending with a soft-max layer to determine the network predictions to classify lesions into three classes. We use ReLU as activation functions. The network had approx. 1.3M parameters.
In addition, to further reducing overfitting, we incorporated a dropout layer \cite{dropout} with a probability of 0.5 during training.

\textbf{\textit{Training Procedure}}. The mean value of the training images was subtracted from each image fed into the CNN. For training we used a batch size of 64 with a learning rate of 0.001 for 150 epochs.
We used stochastic gradient descent optimization with Nesterov momentum updates \cite{Nesterov}, where instead of evaluating the gradient at the current position we evaluated it at the ``look-ahead" position which improves the optimization process.

\section{Generating Synthetic Liver Lesions} \label{sec:methods_generation}

The main problem in training the network described above is the lack of a large labeled training dataset. 
To enlarge the training data and improve the classification results in the liver lesion classification task, we augmented the data in two ways: 1) Classic augmentation that includes varieties of known image manipulations on given data examples; 2) Synthesis of new examples which are learned from the data examples using generative models.
We start with an overview of standard data augmentation techniques and then describe our new method of generating synthetic liver lesion images using generative adversarial networks (GANs).

\subsection{Classic Data Augmentation} \label{sec:data_aug_classic}
Even a small CNN has thousands of parameters that need to be trained. When using deep networks with multiple layers or dealing  with limited numbers of training images, there is a danger of overfitting.
The standard solution to reduce overfitting is data augmentation that artificially enlarges the dataset \cite{Alexnet}.
Classic augmentation techniques on gray-scale images include mostly affine transformations such as translation, rotation, scaling, flipping and shearing \cite{Roth2016,Setio2016}. 
In order to preserve the liver lesion characteristics we avoided transformations that cause shape deformation (like shearing). In addition, we kept the ROI centered around the lesion.

\begin{figure}[!t]
\centering
\includegraphics[width=2.1in]{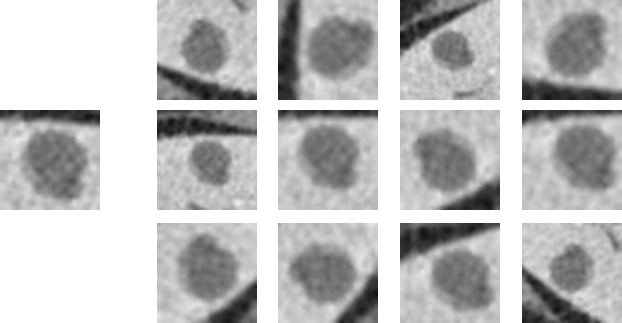}
\caption{Lesion ROI and augmentation examples of translation, rotation, flipping and scaling.}
\label{fig:ROI_augmentation}
\end{figure}

Each lesion ROI was first rotated $N_{rot}$ times at random angles $\theta=[0^{\circ},...,180^{\circ}]$. Afterwards, each rotated ROI was flipped $N_{flip}$ times (up-down,left-right), translated $N_{trans}$ times where we sampled random pairs of $[x,y]$ pixel values between $(-p,p)$ related to the lesion diameter (d) by $p=min(4,0.1\times{d})$. Finally the ROI was scaled $N_{scale}$ times from a stochastic range of scales $s=[0.1\times{d},0.4\times{d}]$. The scale was implemented by changing the amount of tissue around the lesion in the ROI. As a result of the augmentation process, the total number of augmentations was $N = N_{rot} \times{(1+N_{flip}+N_{trans}+N_{scale})}$. An example  lesion and its corresponding augmentations are shown in  Figure \ref{fig:ROI_augmentation}. 
All the ROIs were resized to fit a  uniform size of $64\times{64}$ pixels using bicubic interpolation.

\subsection{Generative Adversarial Networks for Lesion Synthesis} \label{sec:DCGAN}
\begin{figure}[!t]
\centering
\subfloat[]{\includegraphics[width=1.5in]{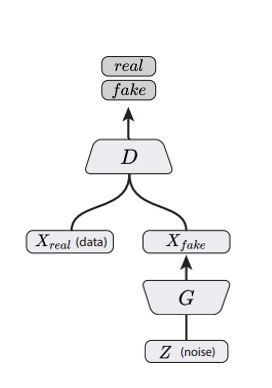}%
\label{fig:DCGAN_arch}}
\hfil
\subfloat[]{\includegraphics[width=1.5in]{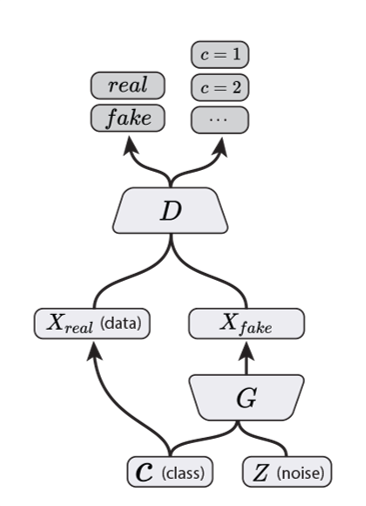}%
\label{fig:ACGAN_arch}}
\caption{(a) DCGAN architecture. (b) ACGAN architecture (Figure is taken from \cite{Odena2016ACGAN}).}
\label{fig:DCGAN_vs_ACGAN_arch}
\end{figure}

GANs \cite{Goodfellow2014GAN} are a specific framework of a generative model. The generative model aims to implicitly learn the data distribution $p_{data}$ from a set of samples ${x^{(1)},...,x^{(m)}}$ (e.g. images) to further generate new samples drawn from the learned distribution.
We explored two variants of GANs for synthesizing labeled lesions, as shown in Figure \ref{fig:DCGAN_vs_ACGAN_arch}: one that generates labeled examples for each lesion class separately and the other that incorporates class conditioning to generate labeled examples all at once.

We started with the first GAN variant, the Deep Convolutional GAN (DCGAN). We followed the architecture proposed by Radford et al.\cite{Radford2015DCGAN}, where both the G and D networks are deep CNNs. They suggested architectural guidelines for stable GAN training and modifications of the original GAN proposed by Goodfellow et al.\cite{Goodfellow2014GAN}, which have become the basis for many recent GAN papers \cite{Salimans2016improvedGAN,Odena2016ACGAN,Yeh2016GANinpainting}.
The model consists of two neural networks that are trained simultaneously (see Figure \ref{fig:DCGAN_arch}). The first network is termed the discriminator and is denoted D. The role of the  discriminator is to discriminate between the real and fake samples. It is inputted a sample $x$ and outputs $D(x)$, its probability of being a real sample.
The second network is termed the generator and is denoted G. The generator synthesizes samples that D will consider to be real samples with high probability. G gets input samples ${z^{(1)},...,z^{(m)}}$ from a known simple distribution $p_{z}$, usually a uniform distribution, and maps $G(z)$ to the image space of distribution $p_{g}$. The goal of G is to achieve $p_{g} = p_{data}$.

Adversarial networks are trained by optimizing the following loss function of a two-player minimax game:
\begin{equation}
\min_{G}\max_{D} \mathbb{E}_{x\sim{p_{data}}} \log{D(x)} + \mathbb{E}_{z\sim{p_{z}}} [\log{(1 - D(G(z)))}]
\label{equation:minimax}
\end{equation} 
The discriminator is trained to maximize $D(x)$ for images with $x\sim{p_{data}}$ and to minimize $D(x)$ for images with  $x\nsim{p_{data}}$.
The generator produces images $G(z)$ to fool D during training such that $D(G(z))\sim{p_{data}}$. Therefore, the generator is trained to maximize $D(G(z))$, or equivalently minimize $1 - D(G(z))$. During training the generator improves in its ability to synthesize more realistic images while the discriminator improves in its ability to distinguish the real from the synthesized images. Hence the moniker of adversarial training.



\begin{figure}[!t]
\centering
\includegraphics[width=3.2in]{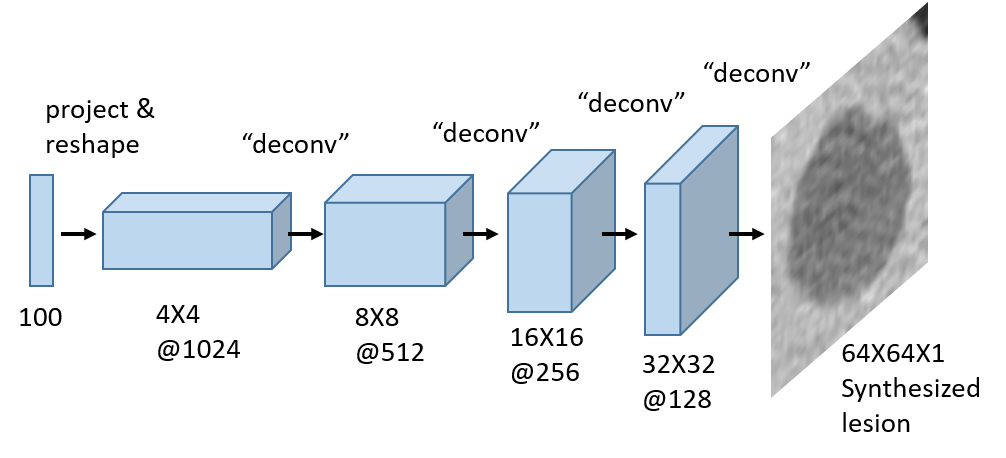}
\caption{Generator architecture (of deep convolutional GAN).}
\label{fig:G_arch_GAN}
\end{figure}

\textbf{\textit{Generator Architecture}}: The generator network takes a vector of 100 random numbers drawn from a uniform distribution as input and outputs a liver lesion image of size $64\times{64}\times{1}$ as shown in Figure \ref{fig:G_arch_GAN}. The network architecture \cite{Radford2015DCGAN} consists of a fully connected layer reshaped to size $4\times{4}\times{1024}$ and four \textit{fractionally-strided convolutional} layers to up-sample the image with a $5\times{5}$ kernel size.  A fractionally-strided convolution (known also as `deconvolution') can be interpreted as expanding the pixels by inserting zeros in between them. Convolution over the expanded image will result in a larger output image. \textit{Batch-normalization} is applied to each layer of the network, except for the output layer. Normalizing responses to have zero mean and unit variance over the entire mini-batch stabilizes the GAN learning process and prevents the generator from collapsing all samples to a single point \cite{batch_normalization}. ReLU activation functions are applied to all layers except the output layer which uses a tanh activation function.

\textbf{\textit{Discriminator Architecture}}: The discriminator network has a typical CNN architecture that takes the input image of size $64\times{64}\times{1}$ (lesion ROI), and outputs one decision: is this lesion real or fake? The network consists of four convolution layers with a kernel size of $5\times{5}$ and a fully connected layer. \textit{Strided convolutions} are applied to each convolution layer to reduce spatial dimensionality instead of using pooling layers. Batch-normalization is applied to each layer of the network, except for the input and output layers. \textit{Leaky ReLU} activation functions $f(x) = \max{(x,leak\times{x})}$ are applied to all layers except the output layer which uses the Sigmoid function for the likelihood probability $(0,1)$ score of the image.

\textbf{\textit{Training Procedure}}: \label{sec:DCGAN_training}
We trained the DCGAN to synthesize liver lesion ROIs for each lesion category separately. The training process was done iteratively for the generator and the discriminator. We used mini-batches of m=64 lesion ROI examples ${x_{l}^{(1)},...,x_{l}^{(m)}}$ for each lesion type $l\in{(Cyst,Metastasis,Hemangioma)}$ and m=64 noise samples ${z^{(1)},...,z^{(m)}}$ drawn from uniform distribution between $[-1,1]$.
The only preprocessing steps used involved scaling the training images to the range of the tanh activation function $(-1,1)$. In the Leaky ReLU, the slope of the leak was set to $leak=0.2$. 
Weights were initialized to a zero-centered normal distribution with standard deviation of 0.02.
We applied stochastic gradient descent with the Adam optimizer \cite{Adam}, an adaptive moment estimation that incorporates the first and second moments of the gradients, controlled by parameters $\beta_{1}=0.5$ and $\beta_{2}=0.999$ respectively. We used a learning rate of 0.0002 for 70 epochs.

\subsection{Conditional Lesion Synthesis} \label{sec:ACGAN}
The second GAN variant is the Auxiliary Classifier GAN (ACGAN).
Conditional GANs are an extension of the GAN model, that enable the model to be conditioned on external information to improve the quality of the generated samples. GAN architectures that incorporate the class labels to produce labeled samples were introduced by \cite{Mirza2014conditionalGAN,Odena2016ACGAN,Salimans2016improvedGAN}. Odena et al.\cite{Odena2016ACGAN} suggested that instead of feeding the discriminator with side information \cite{Mirza2014conditionalGAN}, the discriminator should be tasked with reconstructing side information. This is done by modifying the discriminator to contain an auxiliary decoder network that outputs the class label in addition to the real or fake decision (see Figure \ref{fig:ACGAN_arch}). We followed the architecture proposed in \cite{Odena2016ACGAN} with minor modifications for synthesizing the labeled lesions of all three types.
ACGANs generator architecture is similar to the DCGANs architecture described in section \ref{sec:DCGAN} with class embedding in addition to the input noise samples. The ACGAN discriminator architecture modified the DCGAN to have kernels of size $3\times{3}$ with strided convolutions every odd layer and incorporates a dropout of 0.5 in every layer except for the last layer. We use the ACGAN discriminator without these modifications after optimizing for our small dataset. The discriminator auxiliary decoder classified the three classes of lesions. 

\begin{figure*}[!h]
\centering
\subfloat[]{\includegraphics[width=1.7in]{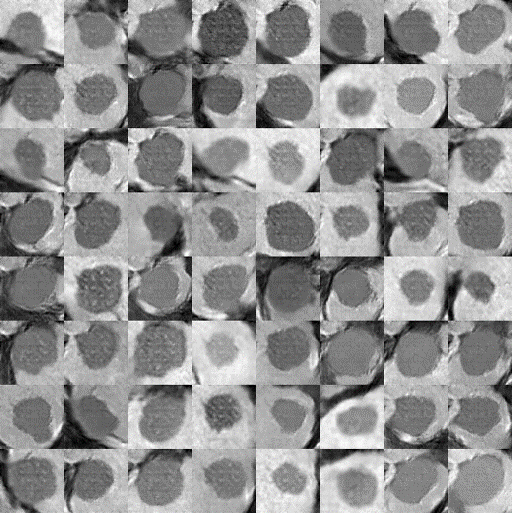}%
\label{fig:synth_cyst}}
\hfil
\subfloat[]{\includegraphics[width=1.7in]{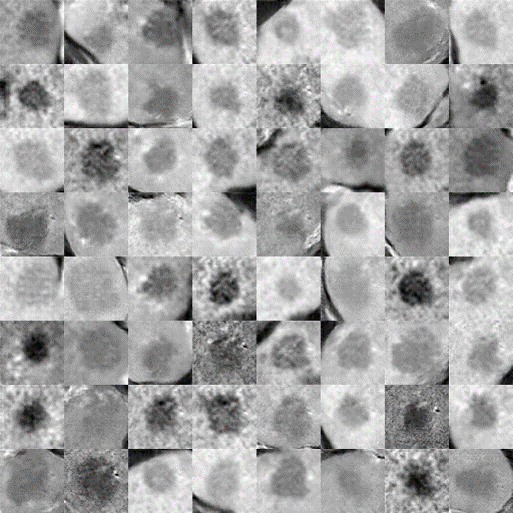}%
\label{fig:synth_met}}
\hfil
\subfloat[]{\includegraphics[width=1.7in]{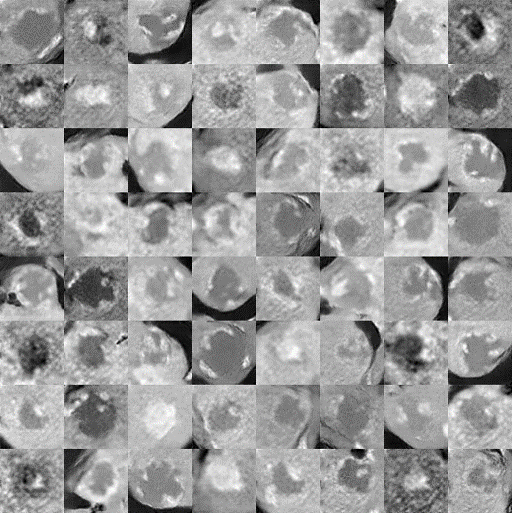}%
\label{fig:synth_hem}}
\caption{Synthetic liver lesion ROIs generated with DCGAN for each category: (a) Cyst examples (b) Metastasis examples (c) Hemangioma examples.}
\label{fig:synth_examples_DCGAN}
\end{figure*}

\textbf{\textit{Training Procedure}}: \label{sec:ACGAN_training}
The training parameters were similar to the ones described in \ref{sec:DCGAN_training} except that we used a learning rate of 0.0001 for 50 epochs. Our training inputs included liver lesion ROIs and their corresponding labels  ${(x_{l},y_{l})^{(1)},...,(x_{l},y_{l})^{(m)}}$ for all lesion types $l\in{(Cyst,Metastasis,Hemangioma)}$, and noise samples ${z^{(1)},...,z^{(m)}}$ drawn from uniform distribution between $[-1,1]$. The loss function needed to be modified to incorporate the label information. For simplification, let us write the basic GAN discriminator maximization equation over the log-likelihood (similar to Equation \ref{equation:minimax}) as:
\begin{equation*}
L=\mathbb{E}[\log{P(S = real|X_{real})]}+\mathbb{E}[\log{P(S=fake|X_{fake})]}
\end{equation*} 
\label{equation:simple_gan_loss} where $P(S|X) = D(X)$ and $X_{fake}=G(z)$. The generator is trained to minimize that objective.
In ACGAN, the discriminator outputs $P(S|X),P(C|X) = D(X)$, and $X_{fake}=G(c,z)$ where $C$ is the class label. The loss has two parts:
\begin{gather*} 
L_{s}=\mathbb{E}[\log{P(S=real|X_{real})]}+\mathbb{E}[\log{P(S=fake|X_{fake})]} \\
L_{c}=\mathbb{E}[\log{P(C=c|X_{real})]}+\mathbb{E}[\log{P(C=c|X_{fake})]}
\end{gather*} 
The discriminator is trained to maximize $L_{s}+L_{c}$ and the generator is trained to maximize $L_{c}-L_{s}$.\\

\section{Experiments and Results} \label{sec:evaluation_results}
 In the following we present a set of experiments and results.
To test the classification results, we employed the CNN architecture described in Section \ref{sec:classification_cnn}.
We then  analyzed the effects of data augmentation using synthetic liver lesions, as compared to  classical data augmentation methodology. 
We implemented the two methods for synthetic lesion generation, as described in Sections \ref{sec:DCGAN} and \ref{sec:ACGAN}. In our experimentations we  found that  the Deep Convolutional GAN (DCGAN) method performed better. We therefore focus on that method in the results presented below. A comparison between the ACGAN and the DCGAN results will be presented in Section 
\ref{Sec:Comparison}.

\subsection{Dataset Evaluation and Implementation Details}
In all experiments and evaluations we used 3-fold cross validation with case separation at the patient level. The number of examples in each fold was $(63,63,62)$ and each contained a balanced number of cyst, metastasis and hemangioma lesion ROIs.
We evaluated the classification performance using a total classification accuracy measure. Additionally, we calculated confusion matrices and sensitivity and specificity measures for each lesion category. All these measures are presented in the following equations: 

\begin{equation}
Total\ Accuracy = \frac{\sum TP}{Amount\ of\ lesions}
\end{equation}

\begin{equation}
Sensitivity = \frac{TP}{TP + FN}
\end{equation}

\begin{equation}
Specificity = \frac{TN}{TN + FP}
\end{equation}

where for each lesion category, positives (P) are examples from this category and negatives (N) are examples from the other two categories.

For the implementation of the liver lesion classification CNN we used the Keras framework \cite{keras}. For the implementation of the GAN architectures we used the TensorFlow framework \cite{tensorflow}. All training processes were performed using an NVIDIA GeForce GTX 980 Ti GPU.

\begin{figure}[!t]
\centering
\includegraphics[width=3.5in]{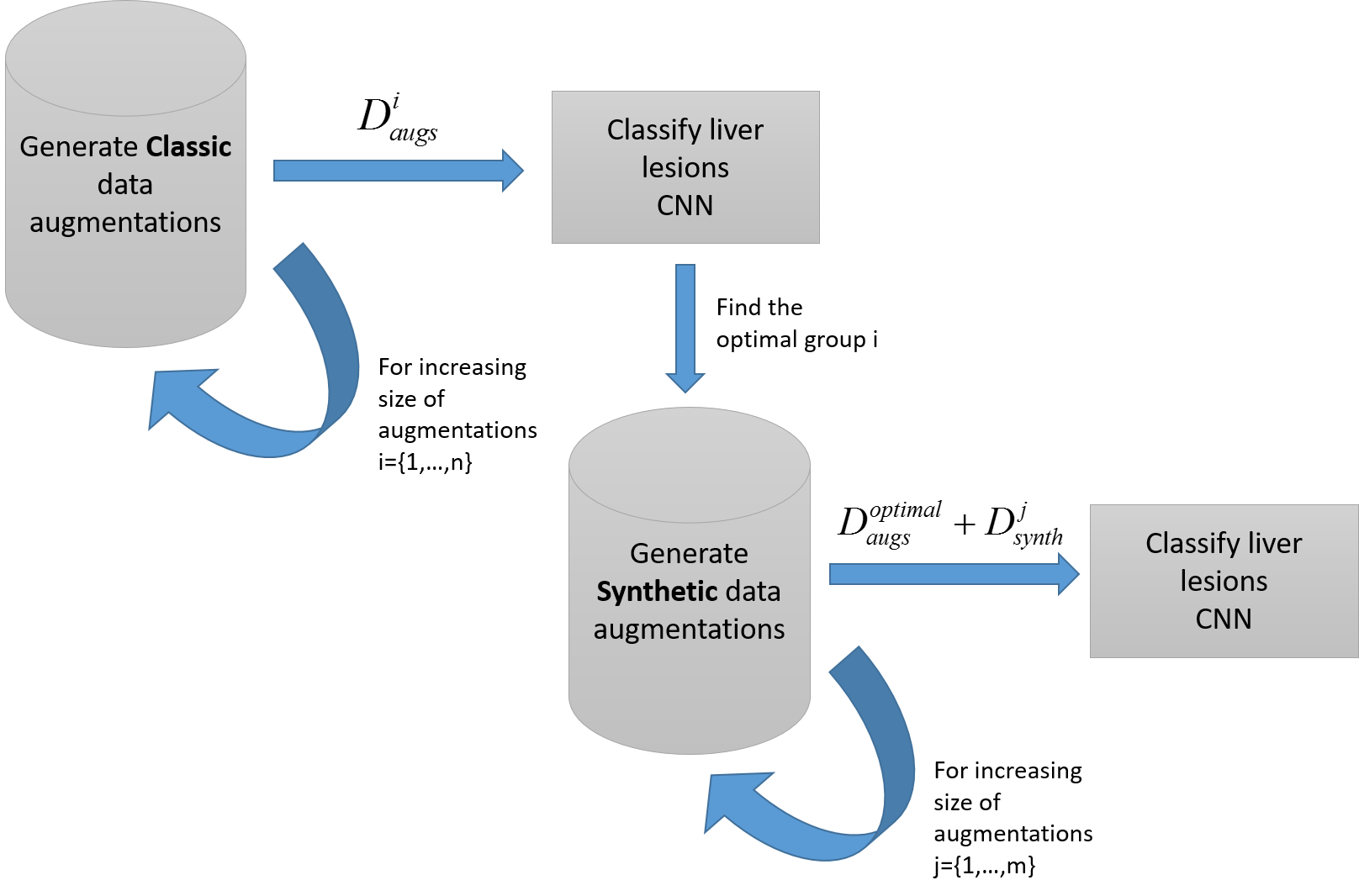}
\caption{Experiment flowchart for evaluating synthetic data augmentation
in the task of classifying liver lesion ROIs.}
\label{fig:experiment_scheme}
\end{figure}

\subsection{Evaluation of the Synthetic Data Augmentation} \label{sec:experiment}
Figure \ref{fig:experiment_scheme} presents the  flowchart for the experiment conducted to evaluate  the results from synthetic data augmentation: 
We started by examining the effects of using only classic data augmentation for the liver lesion classification task (our baseline). We then synthesized liver lesion ROIs using GAN and examined the classification results after adding the synthesized lesion ROIs to the training set. A detailed description of each step is provided next.

\begin{figure*}[h]
\centering
\includegraphics[width=3.7in]{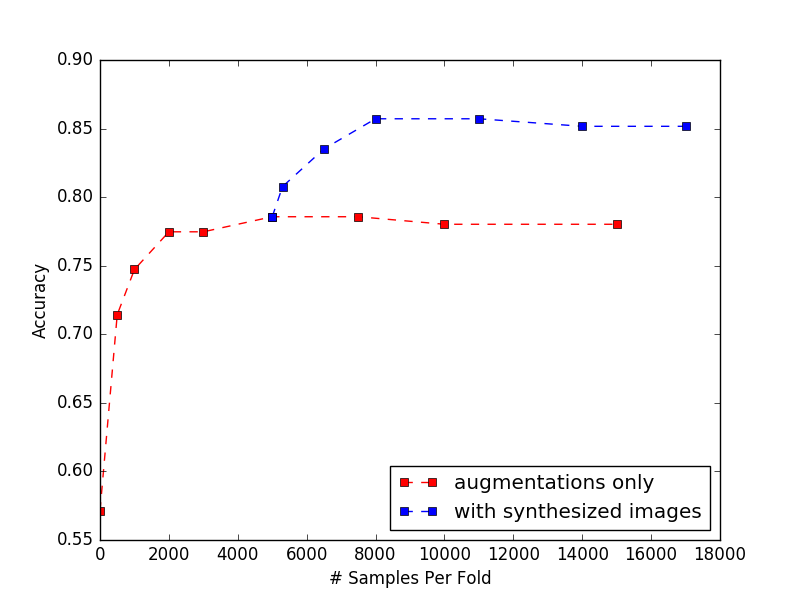}
\caption{Total accuracy results for liver lesion classification of cysts, metastases and hemangiomas with the increase of training set size. The red line shows the effect of adding classic data augmentation and the blue line shows the effect of adding synthetic data augmentation.} 
\label{fig:main_results_graph}
\end{figure*}

\subsubsection{Classical data augmentation}
As our baseline, we used classical data augmentation (see section \ref{sec:data_aug_classic}). We refer to this network 
as CNN-AUG. We recorded  the classification results for the liver lesion classification CNN for increasing amounts of data augmentation over the original training set.
We trained the network and evaluated the results separately for each set of data images (that included the original images and additional classic augmented images), as follows: Let $\{D_{aug}\}_{i=1}^{9}$ be the data groups that include increasing amounts of augmented examples for each training. During testing time, we used the same data examples for all evaluations. 
In order to examine the effect of adding increasing numbers of examples, we formed the data groups additively such that $D_{aug}^{1} \subset D_{aug}^{2} \subset ... \subset D_{aug}^{9}$. The first data group was only made up of the original ROIs. For each original ROI, we produced a large number of augmentations ($N_{rot}=30$, $N_{flip}=3$, $N_{trans}=7$ and $N_{scale}=5$), resulting in $N=480$ augmented images per lesion ROI and overall $\sim 30,000$ examples per folder. Then, we selected the images for the data groups by sampling randomly augmented examples such that for each original lesion we sampled the same augmentation volume. To summarize the augmentation data group preparation process, the number of samples added to each fold (in our 3-folds) was $\{0,500,1000,2000,3000,5000,7500,10000,15000\}$. 
The training process was conducted by cross-validation over 3-folds, such that for each training group, the training examples were from two folds.

\subsubsection{Synthetic data augmentation} \label{sec:experiment_DCGAN_synth}

The second step of the experiment consisted of generating synthetic liver lesion ROIs for data augmentation using GAN. We refer to this network 
as CNN-AUG-GAN. We took the optimal point for the classic augmentation $D_{aug}^{optimal}$ and used this group of data to train the GAN. Since our dataset was too small for effective training, we incorporated classic augmentation for the training process.
We employed the DCGAN architecture 
to train each lesion class separately, using the same 3-fold cross validation process and the same data partition. 
After the generator had learned each lesion class data distribution separately, it was able to synthesize new examples by using an input vector of normal distributed samples (``noise").
Figure \ref{fig:synth_examples_DCGAN} presents examples of synthesized liver lesion ROIs from each class.
The same approach that was applied in step one of the experiment when constructing the data groups was also applied in step two: We collected large numbers of synthetic lesions for all three lesion classes, and constructed data groups $\{D_{synth}\}_{j=1}^{6}$ of synthetic examples additively. To keep the classes balanced, we sampled the same number of synthetic ROIs for each class.
To summarize the synthetic augmentation data group preparation process, the number of samples added to each fold (in our 3-folds) was $\{100\times{3},500\times{3},1000\times{3},2000\times{3},3000\times{3},4000\times{3}\}$. \\

\begin{table}[!ht]
\renewcommand{\arraystretch}{1.3}
\caption{Confusion Matrix for the Optimal Classical Data 
Augmentation Group (CNN-AUG)}
\label{tabel:cm_optimal_augs}
\centering
\begin{tabular}{|c|c|c|c||c|}
\hline
True $\setminus$ Auto & Cyst & Met & Hem & Sensitivity   \\
\hline
Cyst & 52 & 1 & 0 & 98.1\%   \\
\hline
Met & 2 & 44 & 18 & 68.7\%   \\
\hline
Hem & 0 & 18 & 47 & 72.3\%   \\
\hline \hline
Specificity & 98.4\% & 83.9\% & 84.6\% &    \\
\hline
\end{tabular}
\end{table}

\begin{table}[!ht]
\renewcommand{\arraystretch}{1.3}
\caption{Confusion Matrix for the Optimal Synthetic Data Augmentation Group (CNN-AUG-GAN)}
\label{tabel:cm_synth_augs}
\centering
\begin{tabular}{|c|c|c|c||c|}
\hline
True $\setminus$ Auto & Cyst & Met & Hem & Sensitivity   \\
\hline
Cyst & 53 & 0 & 0 & 100\%   \\
\hline
Met & 2 & 52 & 10 & 81.2\%   \\
\hline
Hem & 1 & 13 & 51 & 78.5\%   \\
\hline \hline
Specificity & 97.7\% & 89\% & 91.4\% &    \\
\hline
\end{tabular}
\end{table}

Results of the GAN-based synthetic augmentation experiment are shown in Figure \ref{fig:main_results_graph}.
The baseline results (classical augmentation) are shown in red. We see the total accuracy results for the lesion classification task, for each group of data. When no augmentations were applied, a result of 57\% was achieved; this may be due   to overfitting over the small number of training examples ($\sim 63$ samples per fold). The results improved as the number of training examples increased, up to saturation around 78.6\% where adding more augmented data examples failed to improve the classification results.
We note that the saturation starts with $D_{aug}^{6}=5000$ samples per fold. We define this point as i=optimal where the smallest number of augmented samples were used.
The confusion matrix for the optimal point appears in Table \ref{tabel:cm_optimal_augs}. \\
The blue line in Figure \ref{fig:main_results_graph} shows the total accuracy results for the lesion classification task for the synthetic data augmentation scenario. The classification results improved from 78.6\% with no synthesized lesions to 85.7\% for $D_{aug}^{optimal} + D_{synth}^{3}=5000 + 3000 = 8000$ samples per fold. 
The confusion matrix for the best classification results using synthetic data augmentation is presented in Table \ref{tabel:cm_synth_augs}.

\begin{figure*}[!ht]
\centering
\subfloat[]{\includegraphics[width=2.5in]{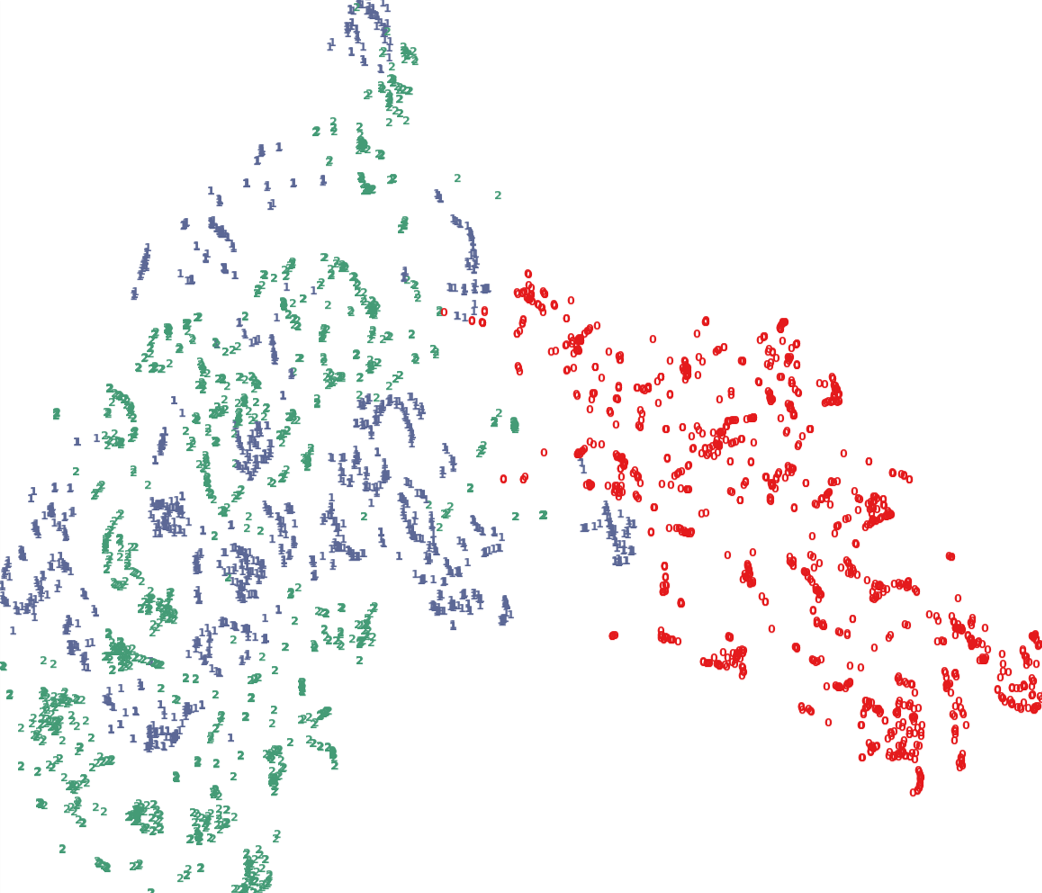}%
\label{fig:tsne_augs}}
\hfil
\subfloat[]{\includegraphics[width=2.5in]{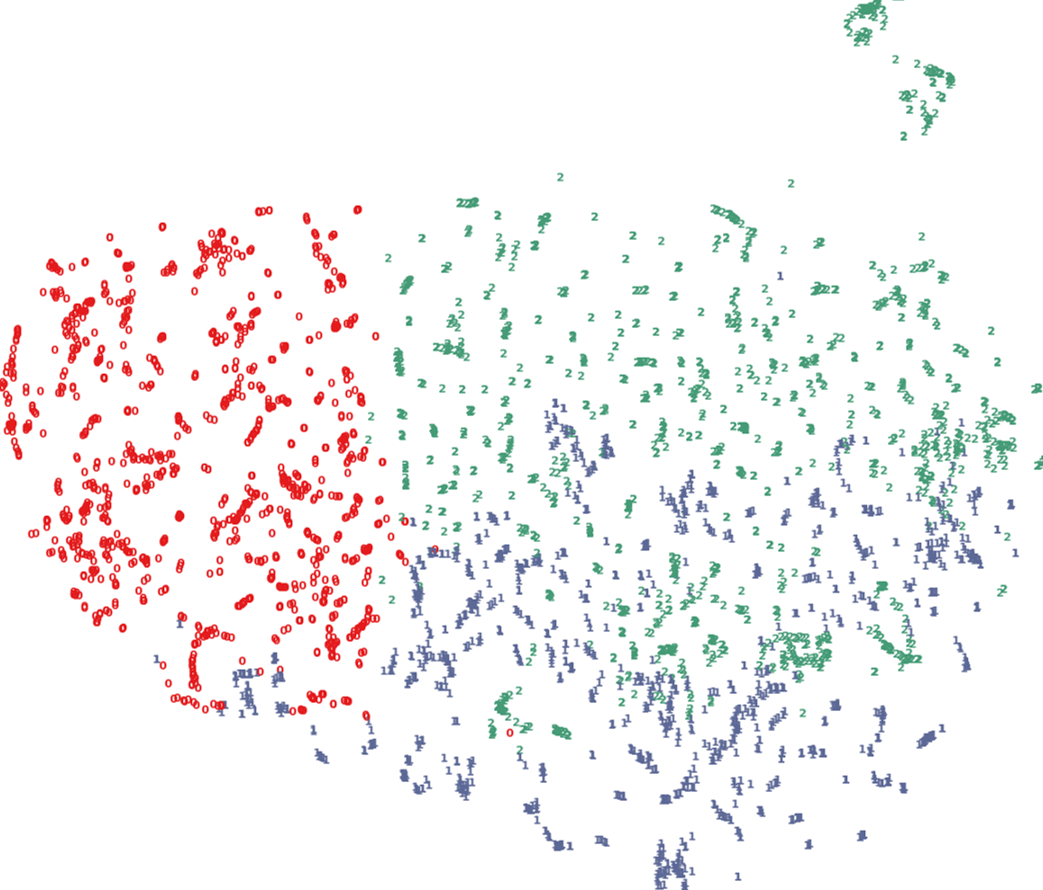}%
\label{fig:tsne_synth}}
\caption{T-SNE embedding of Cysts (red), Metastases (blue) and Hemangiomas (green) real lesion ROIs. (a) Features extracted from CNN-AUG (b) Features extracted from CNN-AUG-GAN.}
\label{fig:tsne_visualization}
\end{figure*}

\subsection{Visualization using t-SNE} 
\label{sec:t-SNE}
To further analyze the results, we used the t-SNE visualization. The t-SNE algorithm for dimensionality reduction enables the embedding of high-dimensional data into a two dimensional space \cite{Maaten2008tsne}. The high-dimensional data for visualization are features extracted from the last layer of a trained liver lesion classification CNN. We trained the CNN in two scenarios: one with the classic augmented data examples (CNN-AUG) and one with the synthesized data examples (CNN-AUG-GAN). Afterwards, for each scenario, we extract the features of real images from the test set and their classic augmentations.
We then used the t-SNE to illustrate the features, as shown in Figure \ref{fig:tsne_visualization} (a) and (b), respectively. 

We note that the cyst category, shown in red, shows a more distinct localization in the t-SNE space. This characteristic correlates well with the more distinctive features of the cyst class as compared to  metastases or hemangiomas. Metastases and hemangiomas have confusing features, which is indicated here in the perceived overlap and accounts for the lower sensitivity and specificity results than in the cyst class. 
When using the synthetic data augmentation, the t-SNE visualization exhibited in general better separating power. This can provide intuition for  the increase in classification performance.

\subsection{Expert Assessment of Synthetic Data} 
Human annotators have been shown to evaluate the visual quality of samples generated by GANs \cite{Salimans2016improvedGAN,Denton2015lapgan}. In our study, we were interested to explore two key points: Is the synthesized lesions appearance a realistic one?  Is the set of lesions generated sufficiently distinct to enable classification amongst the three lesion categories?
These issues were explored with the help of two expert radiologists.

We created an automatic application
which was presented to two independent radiologists, with two tasks. One task was to classify each presented lesion ROI image into one of three classes: cyst, metastasis or hemangioma. The second task was to distinguish between real lesion images and synthetic lesion images.
The experts were given, in random order, lesion ROIs from the  original dataset of 182 real lesions  and 120 additional synthesized lesions. 
Both our algorithm results and the expert radiologists' results were compared against the ground truth classification.

Table \ref{table:app_results} summarizes the  experts' results. We note the overall low results, on the order of 60\%,  in identifying whether the lesions shown were true or fake. 
 In the lesion categorization task,  Expert 1 and Expert 2 classified correctly in 77.8\% and 69.2\% of the cases, respectively. Overall, the radiologists agreed on the lesion class on 222 out of 302 lesions (73.5\%), with a correct classification of 185 out of 302 lesions. 
In addressing these results, it is important to note that the task we defined was not consistent with existing  clinical workflow. The radiologist is trained to make a decision  based on the entire 3D volume, with support from additional anatomical context, medical history context, and more. Here, we challenged the radiologists to reach a decision based on a single 2-D ROI image. 
 In this scenario,  the baseline CNN solution is similar in performance to the human expert. Using the GAN-based augmentation, an increase of approx 
 7\% is achieved.
 
 As a final note, we observe that for both experts, the classification performances for the real lesions and the synthesized lesions were similar. which suggests that our synthetic generated lesions were meaningful in appearance.


 \begin{table}[!t]
\renewcommand{\arraystretch}{1.3}
\caption{Summary of experts' Assessment of Lesion ROI}
\begin{center}
\begin{tabular}{ m{1.2cm}  >{\centering\arraybackslash} m{0.8cm}  >{\centering\arraybackslash} m{0.8cm} >{\centering\arraybackslash} m{1.8cm}   >{\centering\arraybackslash} m{1.8cm} } 
\hline \hline
  & \multicolumn{3}{c}{Classification Accuracy} & Is ROI Real? \\
  & Real & Synthetic & Total Score &  Total Score \\
\hline
 Expert 1  &  78\%  & 77.5\%  &  235$\setminus$302=77.8\%  & 189$\setminus$302=62.5\%\\ 
  \hline
 Expert 2  &  69.2\%  & 69.2\%  &  209$\setminus$302=69.2\%  &   177$\setminus$302=58.6\%\\ 
  \hline \hline
\end{tabular}
\end{center}
\label{table:app_results}
\end{table}

\begin{table}
\renewcommand{\arraystretch}{1.3}
\caption{Performance Comparison for Liver Lesion Classification Between Generative Models}
\begin{center}
\begin{tabular}{ m{4cm}  >{\centering\arraybackslash} m{1.5cm}   >{\centering\arraybackslash} m{1.5cm} } \hline\hline
  Method & Sensitivity & Specificity \\ \hline
CNN-AUG-GAN (DCGAN)  &  \bf{85.7\%}  &  \bf{92.4\%} \\
CNN-AUG-GAN (ACGAN)  &  81.3\%  &  90.0\% \\
ACGAN  discriminator &  79.1\%  &  88.8\% \\ \hline\hline
\end{tabular}
\end{center}
\label{table:comparisons_gan}
\end{table}

\subsection{Comparison with Other Classification Methods} 
\label{Sec:Comparison}

Table \ref{table:comparisons_gan} compares the best classification results between the DCGAN and ACGAN models. As described above, we used synthetic augmentations generated using the DCGAN for training the classification CNN (CNN-AUG-GAN). Training the classification CNN with synthetic augmentations generated using the ACGAN, yield improved results in comparison of using only classic augmentations, but degraded results in comparison to the DCGAN. The ACGAN discriminator contains an auxiliary classifier. Thus, after training the ACGAN, we can use the learned discriminator as an autonomous component to test directly the test set performance. Using this method resulted in $\sim{2}\%$ decrease in performance.


In our final experiment, we compared our CNN classification results for classic augmentation (CNN-AUG) and synthetic augmentation (CNN-AUG-GAN), to a recently published state-of-the-art liver lesion categorization method, termed  BoVW-MI  \cite{Diamant2017BOVWMI}. The BoVW-MI method is an enhancement of the BoVW model.
It  learns a task-driven dictionary of the most relevant visual words per task using a mutual information measure. 
In order to compare between the approaches, using the datasets of the current work, we ran the BoVW-MI method using the specified  optimized parameters for the liver lesion classification task, as found in \cite{Diamant2017BOVWMI}: A patch size of $11\times{11}$, a word size with a 10 PCA coefficient, a dictionary size of 750 words and a MI threshold of 35\%. We trained the BoVW-MI in 3-fold cross validation using the same lesion partitions.
Table \ref{table:comparisons} compares the sensitivity and specificity results of our best results to the BOVW-MI results.

\begin{table*}[!t]
\renewcommand{\arraystretch}{1.3}
\caption{Performance Comparison for Liver Lesion Classification Between CNN and BOVW-MI}
\begin{center}
\begin{tabular}{ m{2.5cm}  >{\centering\arraybackslash} m{1.5cm}  >{\centering\arraybackslash} m{1.5cm} >{\centering\arraybackslash} m{1.5cm}  >{\centering\arraybackslash} m{1.5cm} >{\centering\arraybackslash} m{1.5cm}  >{\centering\arraybackslash} m{1.5cm}} 
\hline \hline
  & \multicolumn{2}{c}{CNN-AUG-GAN} & \multicolumn{2}{c}{CNN-AUG} & \multicolumn{2}{c}{BOVW-MI} \\
  & Sensitivity & Specificity & Sensitivity & Specificity & Sensitivity & Specificity \\
\hline
 Cysts &  100\%  & 97.7\%  &  98.1\%  &  98.4\%  &  96.3\%  & 96.9\% \\ 
  \hline
 Metastases &  81.2\%  & 89.0\%  &  68.7\%  &  83.9\%  &  75.0\%  & 82.2\% \\ 
  \hline
 Hemangiomas &  78.5\%  & 91.4\%  &  72.3\%  &  84.6\%  & 66.1\%  & 87.2\% \\ 
   \hline
 Weighted Average &  \bf{85.7\%}  & \bf{92.4\%}  &  78.6\%  &  88.4\%  &  78.0\%  &  88.3\% \\ 
  \hline \hline
\end{tabular}
\end{center}
\label{table:comparisons}
\end{table*}

\section{Discussion And Conclusions}

This work focused on generating synthetic medical images with GAN for data augmentation to enlarge small datasets and improve performance on classification tasks using CNN. Our relatively small dataset reflects the size of datasets available to most researchers in the medical imaging community (by contrast to the computer vision community where large scale datasets are available).

We tested our hypothesis that adding synthesized examples would improve classification results. The experimental setup is depicted in Figure \ref{fig:experiment_scheme}.
The experiment was carried out on a limited dataset of three liver lesion categories of cysts, metastases and hemangiomas. Each class has its unique features but there is also considerable intra-variability between classes, mostly for the metastases and hemangiomas. We classified the three categories using a CNN architecture.
In running the experiment, we found that increasing the size of the training data groups with the standard augmentation (translation, rotation, flip, scale), improved training results  up to a certain volume of augmented data, where adding more data did not improve the results (Figure \ref{fig:main_results_graph}). 
Table \ref{tabel:cm_optimal_augs} shows the results for the optimal point achieved using the commonly used classic augmentation.

In the second step of the experiment we used GANs to generate new examples learned from our small dataset.
The best generated liver lesion samples were produced by using the Deep Convolution GAN (DCGAN) for each lesion class separately. Starting from the optimal point where classic augmentation reached saturation, we applied increasing sizes of synthetic data. We saw an  improvement in the classification results from 78.6\% to 85.7\% total accuracy (Figure \ref{fig:main_results_graph}). 
We see increase in the sensitivity and specificity of the metastasis and hemangiomas classes. 
It seems that the synthetic data samples generated from a given dataset distribution, using GAN, can add additional variability to the input dataset (Figure \ref{fig:tsne_visualization}), that in turn leads to better performance.

Evaluations of the quality of the synthesized liver lesions were made by two expert radiologists. 
Although the experiment was not conducted in the regular radiologist working environment, and proved to be a challenging task for them, we find it of interest that 
 both experts had the same classification accuracy results for the real set, as well as the  synthesized lesions set (Table \ref{table:app_results}), indicating to us the validity of the lesion generation process. 

In this study, our goal was to assess to what extent synthesized lesions can improve the performance of another system behind the scenes. Our results show that the synthesized lesions have meaningful visualizations and more importantly meaningful features and can be incorporated into computer aided algorithms.

We tested another generative model that incorporated labels in the training process. Both GANs were trained using supervised learning with liver lesion class labels. The DCGAN trained each lesion class separately while the ACGAN trained all three lesion classes at once. 
In recent computer vision studies \cite{Odena2016ACGAN,Salimans2016improvedGAN}, training a GAN that combines label information improved the visualization quality of samples over GANs that did not utilize the label information to generate samples of many classes together. 
Somewhat surprisingly, we found that for our dataset, challenging the discriminator network to perform two tasks (distinguishing real or fake and classifying lesions into 3 categories), resulted in poor results in comparison the DCGAN model. 
Using synthetic augmentation generated using the ACGAN, we were not able to improve the results over the CNN-AUG-GAN (Table \ref{table:comparisons_gan}).

As a final experiment, we compared the  performance of the CNN - based system which we propose in this work, to non-network state-of-the-art methods for liver lesion classification (Table \ref{table:comparisons}). 
Our suggested CNN architecture for classification that employs classic augmentation performed on a par with the BoVW-MI method \cite{Diamant2017BOVWMI}  with the same ROI input. 
Using synthetic data augmentation in our CNN architecture led to the best performance.
 

There are several limitations to this work. 
One possible extension could be an increase from 2-D to 3-D input volumes, using 3-D analysis CNN. 
We trained separate GANs for each lesion class which increased the training complexity. Investigation of  GAN architectures that generate multi-class samples together would be worthwhile.
The quality of the generated lesion samples could possibly be improved by 
incorporating unlabeled data to improve the GAN learning process \cite{Salimans2016improvedGAN}. Further analysis into  modifications of the training loss to incorporate regularization terms for the L1-norm or L2-norm, can be investigated as well \cite{Yeh2016GANinpainting,Schlegl2017ganAnomaly}. 
In the future, we plan to extend our work to additional medical domains that can benefit from synthesis of lesions for improved training.

In conclusion, we presented a method that uses the generation of synthetic medical images for data augmentation to improve performance on a medical problem with limited data. We demonstrated this technique on a liver lesion classification task and achieved an improvement of $\sim 7\%$ using synthetic augmentation over the classic augmentation.
We introduced a CNN-based architecture for the liver lesion classification task, that achieves state-of-the-art results. 
We believe that other medical problems can benefit from using synthetic augmentation, and that the presented approach can lead to   stronger and more robust radiology support systems. 



\ifCLASSOPTIONcaptionsoff
  \newpage
\fi

\bibliography{bibtex/bib/IEEEabrv.bib,bibtex/bib/IEEEexample.bib}{}
\bibliographystyle{IEEEtran}

\end{document}